\documentclass[10pt,twocolumn,letterpaper]{article}

\usepackage{cvpr}
\usepackage{times}
\usepackage{epsfig}
\usepackage{graphicx}
\usepackage{amsmath}
\usepackage{amssymb}
\usepackage{multirow}
\usepackage{booktabs}
\usepackage{color}
\usepackage{array}
\usepackage{footnote}
\usepackage[T1]{fontenc}

\newcommand{\expnumber}[2]{{#1}\mathrm{e}{#2}}

\usepackage[pagebackref=true,breaklinks=true,letterpaper=true,colorlinks,bookmarks=false]{hyperref}

\cvprfinalcopy % *** Uncomment this line for the final submission

\def\cvprPaperID{2130} % *** Enter the CVPR Paper ID here

\ifcvprfinal\fi
\begin{document}

%%%%%%%%% TITLE
\title{Weakly and Semi Supervised Human Body Part Parsing via Pose-Guided Knowledge Transfer}

\author{Hao-Shu Fang$^{1}$,~~Guansong Lu$^{1}$,~~Xiaolin Fang$^{2}$\footnotemark[1],~~Jianwen Xie$^{3}$,~~Yu-Wing Tai$^{4}$,~~Cewu Lu$^{1}$\footnotemark[2]\\
$^{1}$Shanghai Jiao Tong University, China~~$^{2}$ Zhejiang University, China~~\\
$^{3}$University of California, Los Angeles, USA~~$^{4}$ Tencent YouTu\\
{\tt\small fhaoshu@gmail.com sjtuluguansong@gmail.com fxlfang@gmail.com}\\
{\tt\small jianwen@ucla.edu, yuwingtai@tencent.com lucewu@sjtu.edu.cn}}

\maketitle
\renewcommand{\thefootnote}{\fnsymbol{footnote}}
\footnotetext[1]{This work was done when Xiaolin Fang was an intern at MVIG lab of Shanghai Jiao Tong University.}
\footnotetext[2]{The corresponding author is Cewu Lu, email: lucewu@sjtu.edu.cn. Cewu Lu is also a member of SJTU-SenseTime lab and AI research institute of SJTU.}
%%%%%%%%% ABSTRACT
\begin{abstract}
Human body part parsing, or human semantic part segmentation, is fundamental to many computer vision tasks. In conventional semantic segmentation methods, the ground truth segmentations are provided, and fully convolutional networks (FCN) are trained in an end-to-end scheme. Although these methods have demonstrated impressive results, their performance highly depends on the quantity and quality of training data. In this paper, we present a novel method to generate synthetic human part segmentation data using easily-obtained human keypoint annotations. Our key idea is to exploit the anatomical similarity among human to transfer the parsing results of a person to another person with similar pose. Using these estimated results as additional training data, our semi-supervised model outperforms its strong-supervised counterpart by $\mathbf{6}$ mIOU on the PASCAL-Person-Part dataset~\cite{chen2014detect}, and we achieve state-of-the-art human parsing results. Our approach is general and can be readily extended to other object/animal parsing task assuming that their anatomical similarity can be annotated by keypoints. The proposed model and accompanying source code will be made \textbf{publicly available}.
\end{abstract}
%For a given person, our approach generate a part segmentation prior by averaging the morphed part segmentation results of people that have similar poses in the part segmentation dataset. A refinement network is trained to estimate part segmentation results based on the part segmentation prior and the local image features.

%%%%%%%%% BODY TEXT
\vspace{-3mm}
\section{Introduction}
\begin{figure}[t]
\begin{center}
%\fbox{\rule{0pt}{2in} \rule{0.9\linewidth}{0pt}}
   \includegraphics[width=0.9\linewidth]{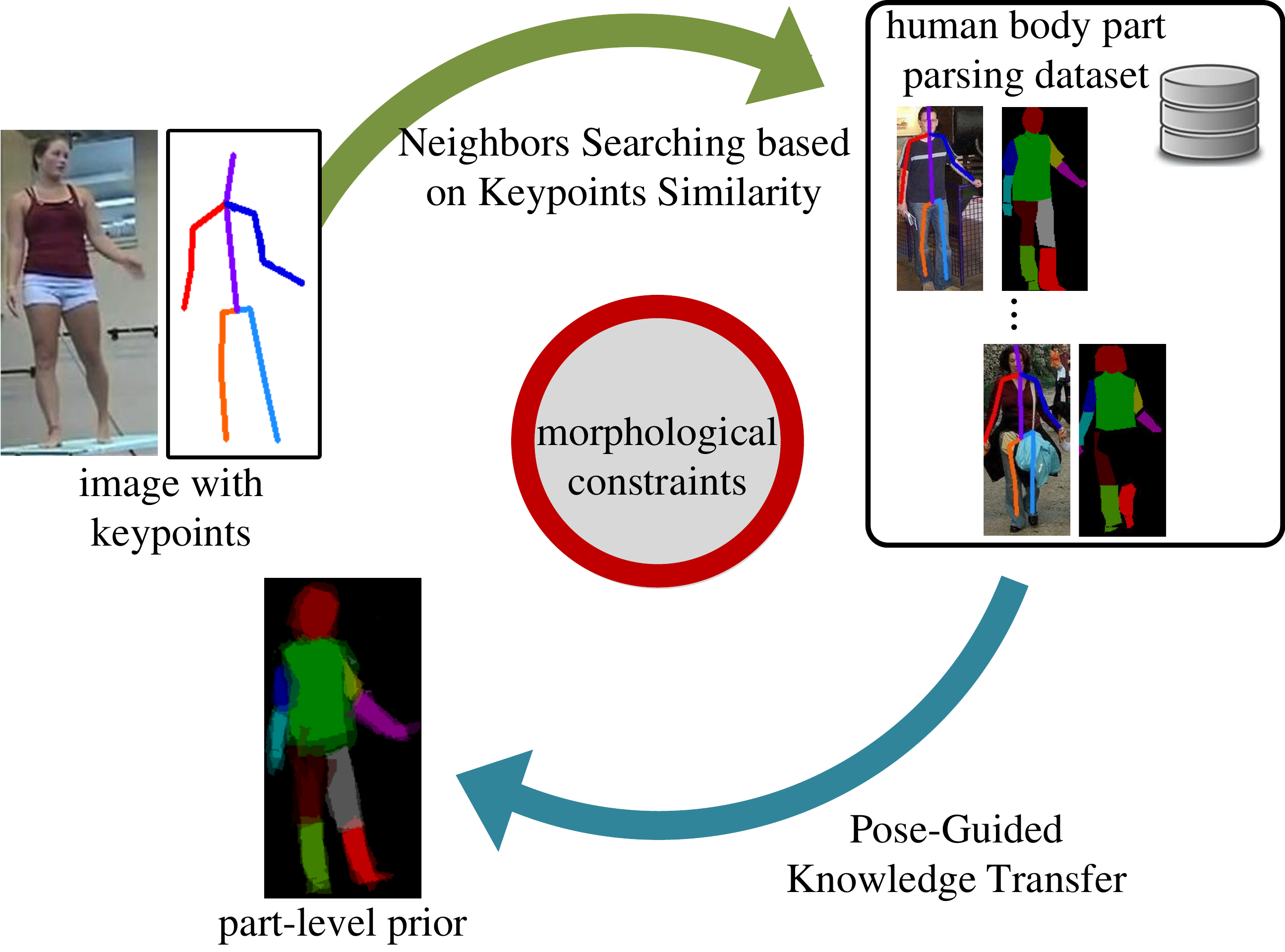}
\end{center}
\vspace{-3mm}
   \caption{Due to the morphological constraints, persons that share the same pose should have similar semantic part segmentations. For a person with only keypoints annotation, we search for persons in the human body part parsing dataset that have similar poses and then transfer their part parsing annotations to the target person. The transferred annotations form a strong part-level prior for the target person.}
\label{fig:intro}
\vspace{-3mm}
\end{figure}

The task of human body part parsing retrieves a semantic segmentation of body parts from the image of a person. Such pixel-level body part segmentations are not only crucial for activity understanding, but might also facilitate various vision tasks such as robotic manipulation~\cite{ekvall2004interactive}, affordances reasoning~\cite{koppula2013learning} and recognizing human-object interactions~\cite{deepparts}. In recent years, deep learning has been successfully applied to the human part parsing problem~\cite{chen2015deeplab,chen2016attention,xia2016zoom}.

To fully exploit the power of deep convolutional neural networks, large-scale datasets are indispensable~\cite{imagenet_cvpr09}. However, semantic labeling of body parts on a pixel-level is labor intensive. For the task of human body part parsing, the largest dataset~\cite{chen2014detect} contains less than 2,000 labeled images for training, which is order of magnitudes less than the amount of training data in common benchmarks for image classification, semantic segmentation and keypoint estimation~\cite{imagenet_cvpr09,lin2014microsoft,Everingham15}. The small amount of data may lead to over-fitting and degrade the performance in real world scenarios.

On the other hand, an abundance of human keypoints annotations~\cite{andriluka14cvpr} is readily available. Current state-of-the-art pose estimation algorithms~\cite{newell2016stacked} have also performed well in natural scene. The human keypoints encode structural information of the human body and we believe that such high-level human knowledge can be transfered to the task of human body part parsing.

However, despite the availability of keypoints annotations, few methods have investigated to utilize keypoints as augmented training data for human body part parsing. The main problem is that human keypoints are a sparse representation, while the human body part parsing requires an enormous amount of training data with dense pixel-wise annotations. Consequently, a end-to-end method which only relies on keypoint annotations as labels may not achieve high performance.

In this paper, we propose a novel approach to augment training samples for human parsing. Due to physical anatomical constraints, humans that share the same pose will have a similar morphology. As shown in Fig~\ref{fig:intro}, given a person, we can use his/her pose annotation to search for the corresponding parsing results with similar poses. These collected parsing results are then averaged and form a strong part-level prior.  In this way, we can convert the sparse keypoint representation into a dense body part segmentation prior. With the strong part-level prior, we combine it with the input image and forward them through a refinement network to generate an accurate part segmentation result. The generated part segmentation can be used as extra data to train a human parsing network.

We conduct exhaustive experiments on the PASCAL-Part dataset~\cite{chen2014detect}. Our semi supervised method achieves state-of-the-art performance using a simple VGG-16~\cite{simonyan2014very} based network, surpassing the performance of ResNet-101~\cite{he2016deep} based counterpart trained on limited part segmentation annotations. When utilizing a model based on the deeper ResNet-101, our proposed method outperforms the state-of-the-art results by \textbf{3} mAP.

%-------------------------------------------------------------------------
\section{Related work}
This paper is closely related to the following areas: semantic part segmentation, joint pose and body part estimation, and weakly supervised learning.
\vspace{-3mm}
\paragraph{Human body part parsing.} In this subtask of semantic segmentation, fully convolutional network (FCN)~\cite{long2015fully} and its variants~\cite{chen2015deeplab,chen2016attention,xia2016zoom} have demonstrated promising results. In ~\cite{chen2015deeplab}, Chen \emph{et al.} proposed atrous convolution to capture object features at different scales and they further combined the convolutional neural network (CNN) with a Conditional Random Field (CRF) to improve the accuracy. In~\cite{chen2016attention}, the authors proposed an attention mechanism that softly combines the segmentation predictions at different scales according to the context. To tackle the problem of scale and location variance, Xia \emph{et al.}~\cite{xia2016zoom} developed a model that adaptively zoom the input image into the proper scale to refine the parsing results.

Another approach to human parsing is the usage of recurrent networks with long short term memory (LSTM) units\cite{hochreiter1997long}. The LSTM network can innately incorporate local and global spatial dependencies into their feature learning. In \cite{liang2016semantic}, Liang \emph{et al.} proposed the local-global LSTM network to incorporate spatial dependencies at different distances to improve the learning of features. In \cite{liang2016semanticG}, the authors proposed a Graph LSTM network to fully utilize the local structures (e.g., boundaries) of images. Their network takes arbitrary-shaped superpixels as input and propagates information from one superpixel node to all its neighboring superpixel nodes. To further explore the multi-level correlations among image regions, Liang \emph{et al.}~\cite{liang2017interpretable} proposed a structure-evolving LSTM that can learn graph structure during the optimization of LSTM network. These LSTM networks achieved competitive performance on human body part parsing.
\vspace{-3mm}
\paragraph{Utilizing Pose for Human Parsing.} Recent works try to utilize the human pose information to provide high-level structure for human body part parsing. Promising methods include pose-guided human parsing~\cite{xia2016pose}, joint pose estimation and semantic part segmentation~\cite{dong2014towards,ladicky2013human,xia2017joint} and self-supervised structure-sensitive learning~\cite{gong2017look}. These methods focus on using pose information to regularize part segmentation results, and they lie in the area of strong supervision. Our method differs from theirs in that we aim to transfer the part segmentation annotations to unlabeled data based on pose similarity and generate extra training samples, which emphasizes semi-supervision.
\begin{figure*}[hbt]
\begin{center}
\includegraphics[width=0.9\linewidth]{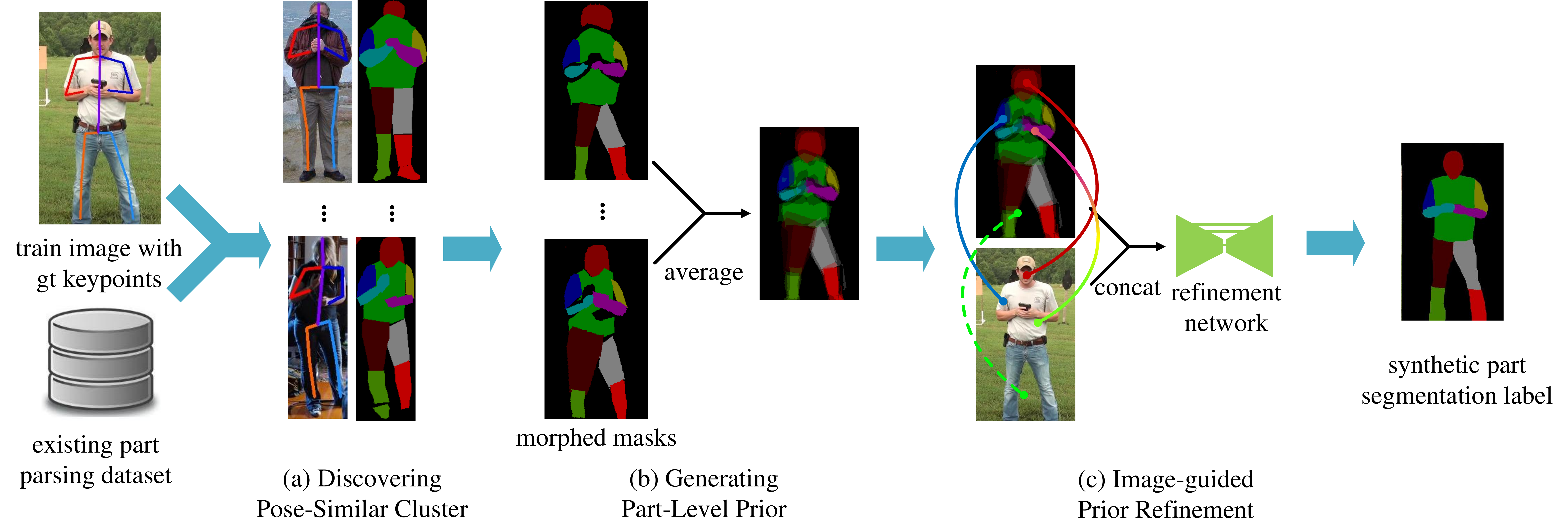}
\end{center}
\vspace{-3mm}
   \caption{Overview of our method. Given an image with keypoint annotations, we first search for the corresponding part segmentation with similar pose (a). Then, we apply a pose-guided morphing to the retrieved part segmentation masks and compute a part-level prior (b). A refinement network is applied to refine the prior based on local image evidence. The semantic part segmentation results can be used as extra training data to train a human parsing network without keypoint annotations (c). See text for more details.}
\label{fig:overview}
\vspace{-3mm}
\end{figure*}
\vspace{-3mm}
\paragraph{Weak Supervision for Semantic Segmentation.}
In \cite{pathak2014fully,chen2014enriching,papandreou2015weakly,dai2015boxsup,lin2016scribblesup,bearman2016s}, the idea is to utilize weakly supervised methods for semantic segmentation. In particular, Chen \emph{et al.}~\cite{chen2014enriching} proposed to learn segmentation priors based on visual subcategories. Dai \emph{et al.}~\cite{dai2015boxsup} harnessed easily obtained bounding box annotations to locate the object and generated candidate segmentation masks using unsupervised region proposal methods. Since bounding box annotations cannot generalize well for background (e.g. water, sky, grasses), Li \emph{et al.}~\cite{lin2016scribblesup} further explored training semantic segmentation models using the sparse scribbles. In~\cite{bearman2016s}, Bearman \emph{et al.} trained a neural network using supervision from a single point of each object. Under some time budget, the proposed supervised method may yield improved result compared to other weak supervised counterparts.

\section{Our Method}
\subsection{Problem Definition}
Our goal is to utilize pose information to weakly supervise the training of a human parsing network. Consider a semantic part segmentation problem where we have a dataset $\mathcal{D}_{s} = \{I_{i}, S_{i}, K_{i}\}_{i=1}^{N}$ of $N$ labeled training examples. Let $I_{i} \in \mathrm{R}^{h\times w \times 3}$ denote an input image, $S_{i} \in \mathrm{R}^{h\times w \times u}$ denotes its corresponding part segmentation label, $K_{i}\in \mathrm{R}^{v\times 2}$ denotes its keypoints annotation, and $i$ is training example index. Each example contains at most $u$ body parts and $v$ keypoints. In practice, $N$ is usually small since labeling the human semantic part is very labor intensive.

For a standard semantic part segmentation problem, the objective function can be written as:
\begin{equation}
\mathcal{E}(\Phi) = \sum_{i} \sum_{j}e[f_{p}^{j}(I_{i};\Phi),S_{i}(j)],
\label{eq:seg}
\end{equation}
where $j$ is the pixel index of the image $I_{i}$, $S_{i}(j)$ is the semantic label at pixel $j$, $f_{p}(\cdot)$ is the fully convolutional network, $f_{p}^{j}(I_{i};\Phi)$ is the per-pixel labeling produced by the network given parameters $\Phi$, and $e[\cdot]$ is the per-pixel loss function.

Similarly, we can consider another dataset ${\mathcal{D}_{p} = \{I_{i}, K_{i}\}_{i=1}^{M}}$ of $M$ examples with only keypoints annotations where $M \gg N$. Our goal is to generate part segmentations for all images in $\mathcal{D}_{p}$ and utilize these as additional training data when training a FCN by minimizing the objective function in Eqn.~\ref{eq:seg}.

\subsection{Overview}
To utilize the keypoints annotations, we directly generate the pixel-wise part segmentations based on keypoints annotations. Given a target image $I_{t}$ and keypoints annotation $K_{t}$ where $(I_{t}, K_{t}) \in \mathcal{D}_{p}$, we first find a subset of images in $\mathcal{D}_{s}$ that have the most similar keypoints annotations (Sec.~\ref{sec:cluster}). Then, for the clustered images, we apply pose-guided morphing to their part segmentations to align them with the target pose (Sec.~\ref{sec:generation}). The aligned body part segmentations are averaged to generate the part-level prior for the target image $I_{t}$. Finally, a refinement network is applied to estimate the body part segmentation of $I_{t}$ (Sec.~\ref{sec:refinement}). The training method of our refinement network will be detailed in Sec.~\ref{sec:refineNet}. The generated part segmentations can be used as extra training data for human part segmentation (Sec.~\ref{sec:semi}).  Figure~\ref{fig:overview} gives an overview of our method.

%-------------------------------------------------------------------------
\subsection{Discovering Pose-Similar Clusters}
\label{sec:cluster}
In order to measure similarity between different poses, we first need to normalize and align all the poses. We normalize the pose size by fixing their torsos to the same length. Then, the hip keypoints are used as the reference coordinate to align with the origin. We measure the Euclidean distances between $K_{t}$ and every keypoint annotations in $\mathcal{D}_{s}$. The top-$k$ persons in $\mathcal{D}_{s}$ with the smallest distances are chosen and form the pose-similar cluster, which serves as the basis for the following part-level prior generation step. The influence of $k$ will be evaluated in Sec.~\ref{para:strategy}.

Intuitively, given an image $I_{t}$ with only keypoint annotations $K_{t}$, one may think of an alternative solution to obtain the part-parsing prior by solely morphing the part segmentation result of the person that has the closest pose to $K_{t}$. However, due to the differences between human bodies or possible occlusions, a part segmentation result with distinct boundary may not fit well to another one. Thus, instead of finding the person with the most similar pose, we find several persons that have similar poses and generate part-level prior by averaging their morphed part segmentation results. Such averaged part-level prior can be regarded as a probability map for each body part. It denotes the possibility for each pixel of whether it belongs to a body part based on real data distribution. In Sec.~\ref{para:strategy}, we show that using the averaged prior achieves better performance than using only the parsing result of the closest neighbor.

\subsection{Generating Part-Level Prior}
\label{sec:generation}
The discovered pose-similar cluster forms a solid basis for generating the part-level prior. However, for each cluster, there are some inevitable intra-cluster pose variations, which makes the part parsing results misaligned. Thus, we introduce the pose-guided morphing method.
\begin{figure}[t]
\begin{center}
%\fbox{\rule{0pt}{2in} \rule{0.9\linewidth}{0pt}}
   \includegraphics[width=1\linewidth]{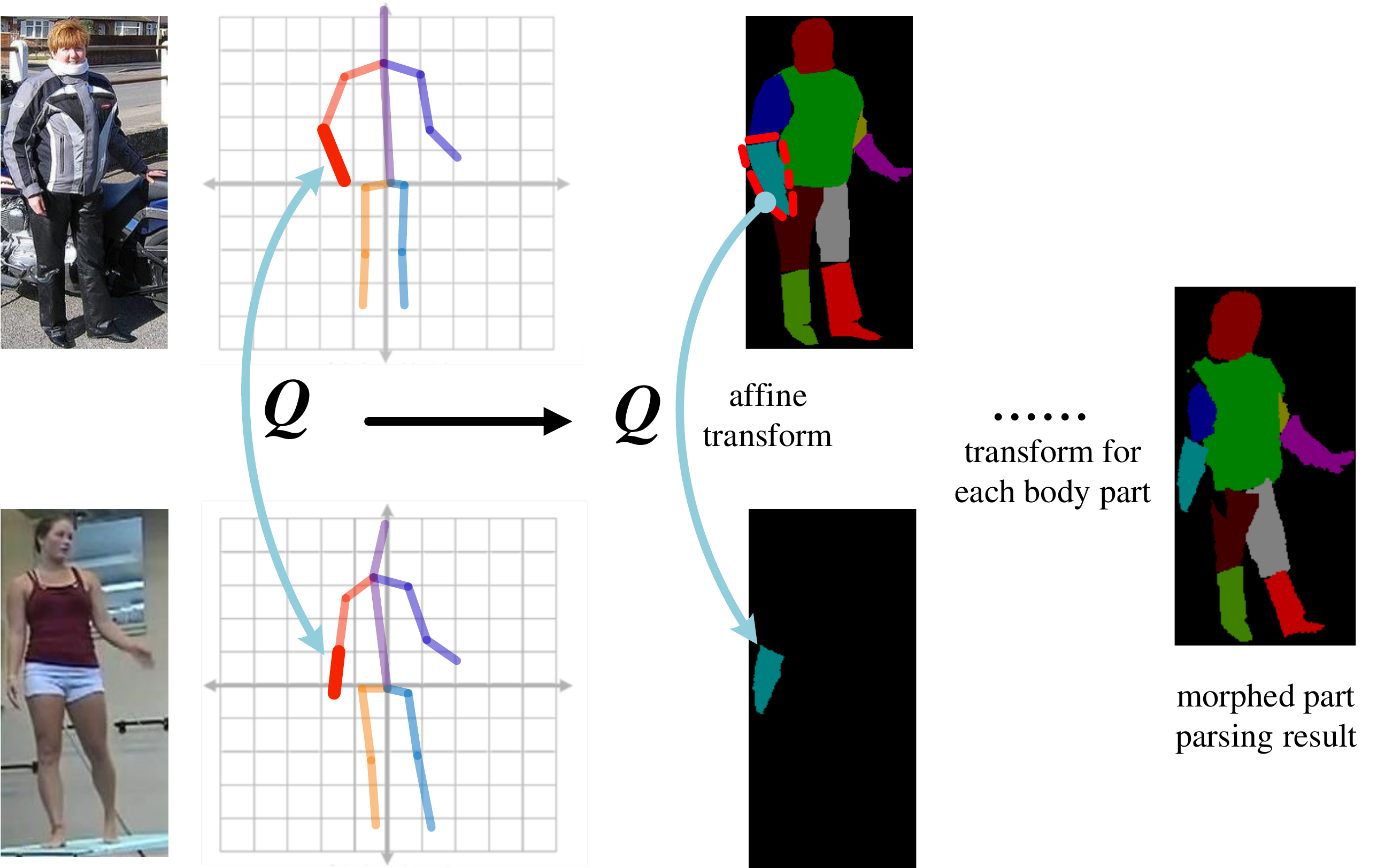}
\end{center}
\vspace{-4mm}
   \caption{Pose-guided morphing for body part parsing. For each body part, we compute an affine transformation matrix \textbf{\emph{Q}} according to the corresponding pose segment. The local body part is then transformed according to the estimated transformation matrix.}
   \vspace{-3mm}
\label{fig:morphing}
\end{figure}

For $n$ persons in the same pose-similar cluster, let us denote their part parsing results as $\mathbb{S} = \{S_{1},...,S_{n}\}$, their keypoints annotations as $\mathbb{K} = \{K_{1},...,K_{n}\}$ and the morphed part parsing results as $\mathbb{\widetilde{S}} = \{\widetilde{S}_{1},...,\widetilde{S}_{n}\}$. By comparing the poses in $\mathbb{K}$ with the target pose $K_{t}$, we can compute the transformation parameters $\theta$ and then use them to transform $\mathbb{S}$ to obtain the morphed part parsing results $\mathbb{\widetilde{S}}$. We use the affine transformation to morph the part segmentations. This procedure can be expressed as:
\begin{equation}
\mathbb{\widetilde{S}} = \{T(S_{i};\theta_{i})~|~1\leq i\leq n~,\,~S_{i}\in \mathbb{S}\},
\end{equation}
where
\begin{equation}
    \theta_{i} = g(K_{t},K_{i}), K_{i}\in \mathbb{K},\nonumber
\end{equation}
$T(\cdot)$ is the affine transformation with parameters $\theta$, and $g(\cdot)$ computes $\theta$ according to pose $K_{i}$ and the target pose $K_{t}$.

For the part parsing annotations, we represent them as the combination of several binary masks. Each mask represent the appearance of a corresponding body part. The morphing procedure is conducted on each body part independently. Consider the left upper arm as an example. For the left upper arm segment $G_{1} = \boldsymbol{\vec{x}_{1}}$ of pose $K_{1}$ and the same segment $G_{2} = \boldsymbol{\vec{x}_{2}}$ of pose $K_{t}$, we have transformation relationship
\begin{equation}\label{eq2}
\left(
\begin{matrix}
   \boldsymbol{\vec{x}_{1}} \\
   1
  \end{matrix}
  \right) =
\mathrm{Q}
\left(
\begin{matrix}
   \boldsymbol{\vec{x}_{2}} \\
   1
  \end{matrix}
\right)
=
\left[
  \begin{matrix}
  \boldsymbol{A} & \boldsymbol{\vec{b}}\\
  \boldsymbol{0} & 1
  \end{matrix}
\right]
\left(
\begin{matrix}
   \boldsymbol{\vec{x}_{2}} \\
   1
  \end{matrix}
\right),
\end{equation}
where $\mathrm{Q}$ is the affine transformation matrix we need to calculate. Since both $\boldsymbol{\vec{x}_{1}}$ and $\boldsymbol{\vec{x}_{2}}$ are known, we can easily compute the result of $\boldsymbol{A}$ and $\boldsymbol{\vec{b}}$. Then, the morphed part segmentation mask can be obtained by
\begin{equation}\label{eq1}
\left(
\begin{matrix}
   x_{i}^{t} \\
   y_{i}^{t}
  \end{matrix}
  \right) =
\boldsymbol{A}
\left(
\begin{matrix}
   x_{i}^{s} \\
   y_{i}^{s}
  \end{matrix}
\right)
+\boldsymbol{\vec{b}},
\end{equation}
where $\{x_{i}^{t},y_{i}^{t}\}$ and $\{x_{i}^{s},y_{i}^{s}\}$ are the coordinates before and after transformation. Figure~\ref{fig:morphing} illustrates our pose-guided morphing method.

After the pose-guided morphing, the transformed part segmentations $\mathbb{\widetilde{S}}$ are averaged to form the part-level prior
\begin{equation}
P_{t} = \frac{1}{n}\sum_{i=1}^{n}\widetilde{S}_{i}.\nonumber
\end{equation}
\subsection{Prior Refinement}
\label{sec:refinement}
Finally, we feed forward our part-level prior through a refinement network together with the original image.

With a coarse prior, the search space for our refinement network is significantly reduced, and thus it can achieve superior results than directly making predictions from a single image input. The discriminative power of a CNN can eliminate the uncertainty at the body part boundary of the part parsing prior based on local image evidence, thus leading to high-quality parsing results. For each image $I_{t}$ with body part prior $P_{t}$, we estimate the part segmentation result $\widehat{S}_{t}$ by
\begin{equation}
\widehat{S}_{t} = f_{r}(I_{t},P_{t};\Psi),
\vspace{-3mm}
\end{equation}
where $f_{r}$ is the refinement network with parameters $\Psi$. In the next section, we will elaborate on the learning of $\Psi$. This estimated part segmentation result can be used as extra training data to train the FCN. The semi-supervised regime will be further discussed in Section~\ref{sec:semi}.

\subsection{Training of Refinement Network}
\label{sec:refineNet}
\begin{figure}[t]
\begin{center}
   \includegraphics[width=0.9\linewidth]{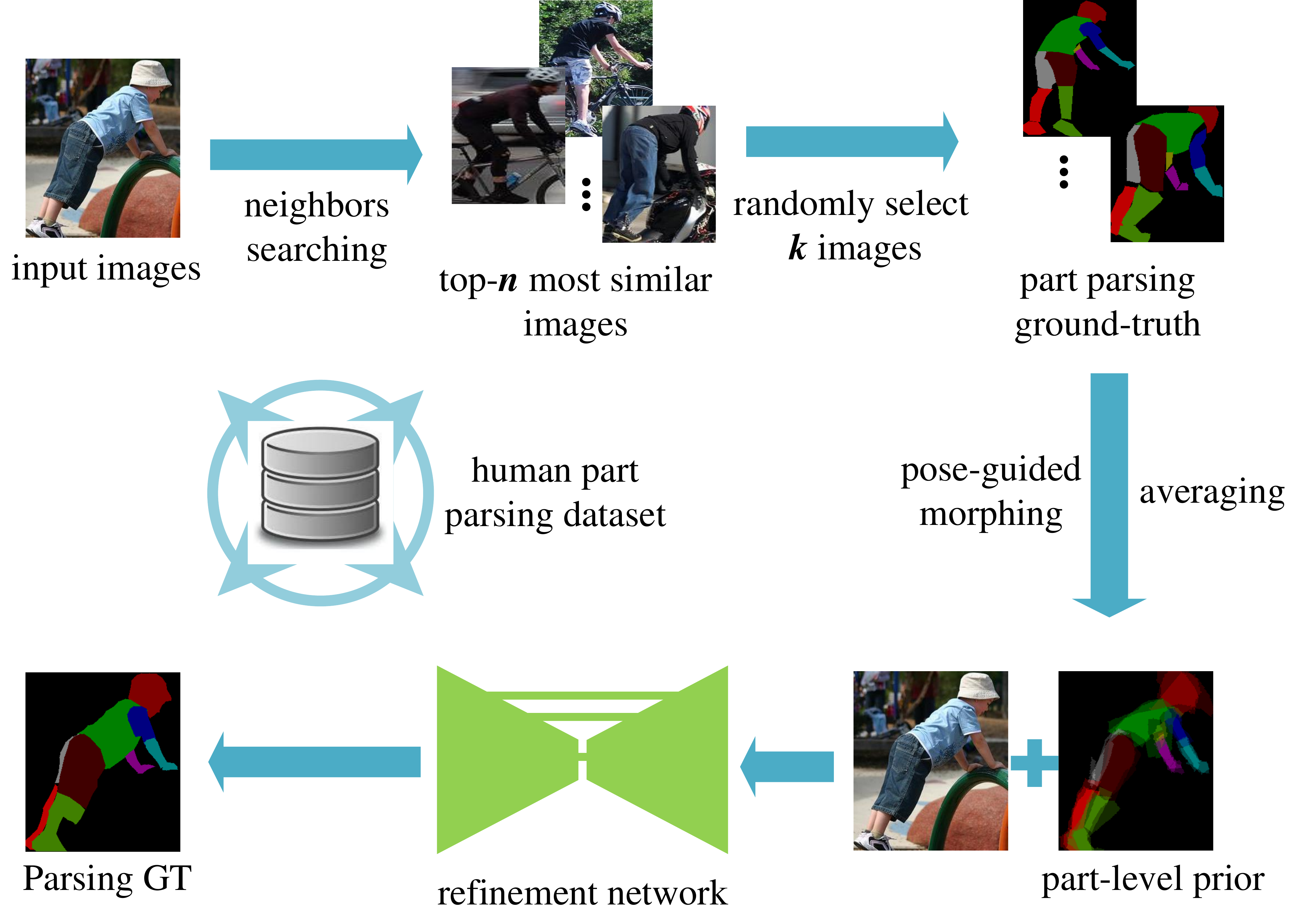}
\end{center}
\vspace{-3mm}
   \caption{Training pipeline for our refinement network. We elaborate on the details in the text.}
\label{fig:refineNet}
\vspace{-3mm}
\end{figure}
Previous sections are conducted under the assumption that we have a well-trained refinement network. In this section, we will explain the training algorithm. Fig.~\ref{fig:refineNet} depicts a schematic overview of the proposed training pipeline.

The refinement network is a variant of ``U-Net'' proposed in~\cite{pix2pix2016}, which is in the form of an auto-encoder network with skip connections. For such a network, the input will be progressively down-sampled until a bottleneck layer and then gradually up-sampled to restore the input size. To preserve the local image evidence, skip connections are introduced between each layer $i$ and layer $n - i$, assuming the network has $n$ layers in total. Each skip connection concatenates the feature maps at layer $i$ with those in layer $n - i$. We refer readers to~\cite{pix2pix2016} for the detailed network structures. The input for our network is an image as well as a set of masks. Each mask is a heatmap ranging from 0 to 1 that represents the probability map for a specific body part. The output for this network is also a set of masks that have the same representation. The label is a set of binary masks indicating the appearance of each body part. Our objective function is the L1 distance between the output and the label. Both input and output size are set as $256\times 256$.

To train the refinement network, we utilize the data in  $\mathcal{D}_{s}$ with both semantic part segmentation annotations and pose annotations. Given an image $I_{m} \in \mathcal{D}_{s}$, similar to the pipeline for generating part-level prior in Sec. \ref{sec:generation}, we first generate the part parsing prior $P_{m}$ given $I_{m}$. The only difference is that when discovering pose-similar cluster, we find $n$ nearest neighbours each time and randomly pick $k$ of them to generate part-level prior. This can be regarded as a kind of data augmentation to improve the generalization ability of the refinement network. The impact of $n$ will be discussed in Sec.~\ref{para:strategy}. Formally, with the part-level prior $P_{m}$ and semantic part segmentation ground truth $S_{m}$ for image $I_{m}$, we train our refinement network by minimizing the cost function:
\begin{equation}
\mathcal{E}(\Psi) = \sum_{j} \lVert S_{m}(j)-f_{r}^{j}(I_{m},P_{m};\Psi) \rVert_{1} .
\vspace{-2mm}
\end{equation}

\subsection{Semi-Supervised Training for Parsing Network}
\vspace{-2mm}
\label{sec:semi}
In previous sections, we have presented our method to generate pixel-wise part segmentation labels based on keypoints annotations. Now we can train the parsing network for part segmentation in a semi-supervised manner. For our parsing network, we use the VGG-16 based model proposed in~\cite{chen2016attention} due to its effective performance and simple structure. In this network, multi-scale inputs are applied to a shared VGG-16 based DeepLab model~\cite{chen2015deeplab} for predictions. A soft attention mechanism is employed to weight the outputs of the FCN over scales. The training for this network follows the standard process of optimizing the per-pixel regression problem, which is formulated as Eqn.~\ref{eq:seg}. For the loss function, we use the multinomial logistic loss. During training, the input image is resized and padded to $320\times 320$.

We consider updating the parameter $\Phi$ of network $f_{p}$ by minimizing the objective function in Eqn.~\ref{eq:seg} on $\mathcal{D}_{s}$ with ground truth labels and $\mathcal{D}_{p}$ with generated part segmentation labels.
\vspace{-2mm}
\section{Experiments}
\vspace{-2mm}
In this section, we first introduce related datasets and implementation details in our experiments. Then, we report our results and comparisons with state-of-the-art performance. Finally, we conduct extensive experiments to validate the effectiveness of our proposed semi-supervision method.
\subsection{Datasets}
\paragraph{Pascal-Person-Part Dataset~\cite{chen2014detect}} is a dataset for human semantic part segmentation. It contains 1,716 images for training and 1,817 images for testing. The dataset contains detailed pixel-wise annotations for body parts, including hands, ears, \emph{etc}. The keypoint annotations for this dataset have been made available by~\cite{xia2017joint}.
\vspace{-3mm}
\paragraph{MPII~\cite{andriluka14cvpr}} is a challenging benchmark for person pose estimation. It contains around 25K images and over 40K people with pose annotations. The training set consists of over 28K people and we select those with full body annotations as extra training data for human part segmentation. After filtering, there remain 10K images with single person pose annotations.
\vspace{-3mm}
\paragraph{Horse-Cow Dataset~\cite{wang2015semantic}} is a part segmentation benchmark for horse and cow images. It contains 294 training images and 227 testing images. The keypoint annotations for this dataset are provided by~\cite{PoseletsPAMI}. In addition, 317 extra keypoint annotations are provided by~\cite{PoseletsPAMI}, which have no corresponding pixel-wise part parsing annotations. We take these keypoint annotations as extra training data for our network.
\subsection{Implementation Details}
\label{sec:implem}
In our experiments, we merge the part segmentation annotations in~\cite{chen2014detect} to be Head, Torso, Left/Right Upper Arms, Left/Right Lower Arms, Left/Right Upper Legs and Left/Right Lower Legs, resulting in 10 body parts. The pose-guided morphing is conducted on the mask of each body part respectively. In order to be consistent with previous works, after the morphing, we merge the masks of each left/right pair by max-pooling and get six body part classes.

For our semi-supervision setting, we first train the refinement network and then fix it to generate synthetic part segmentation data using keypoints annotations in~\cite{andriluka14cvpr}. Note that for simplicity, we only synthesize part segmentation labels for the single person case, and it would easy to extend to a multi-person scenario. To train our refinement network with single person data, we crop those people with at least upper body keypoints annotations in the training list of the part parsing dataset~\cite{chen2014detect} and get 2,004 images with a single person and corresponding part segmentation label. We randomly sample 100 persons as validation set for the refinement network to set hyper-parameters.

To train the refinement network, we apply random jittering by first resizing input images to $286 \times 286$ and then randomly cropping to $256 \times 256$. The batch size is set to 1. We use the Adam optimizer with an initial learning rate of $\expnumber{2}{-4}$ and $\beta_1$ of 0.5. To train our parsing network, we follow the same setting as~\cite{chen2016attention} by setting the learning rate as 1e-3 and decayed by 0.1 after 2,000 iterations. The learning rate for the last layer is 10 times larger than previous layers. The batch size is set to 30. We use the SGD solver with a momentum of 0.9 and weight decay of $\expnumber{5}{-4}$. All experiments are conducted on a single Nvidia Titan X GPU. Our refinement network is trained from scratch for 200 epochs, which takes around 18 hours. The parsing network is initialized with a model pre-trained on COCO dataset~\cite{lin2014microsoft} which is provided by the author. We train it for 15K iterations, and it takes around 30 hours for training.

\subsection{Results and Comparisons}
\begin{figure}[t!]
\begin{center}
\includegraphics[width=1\linewidth]{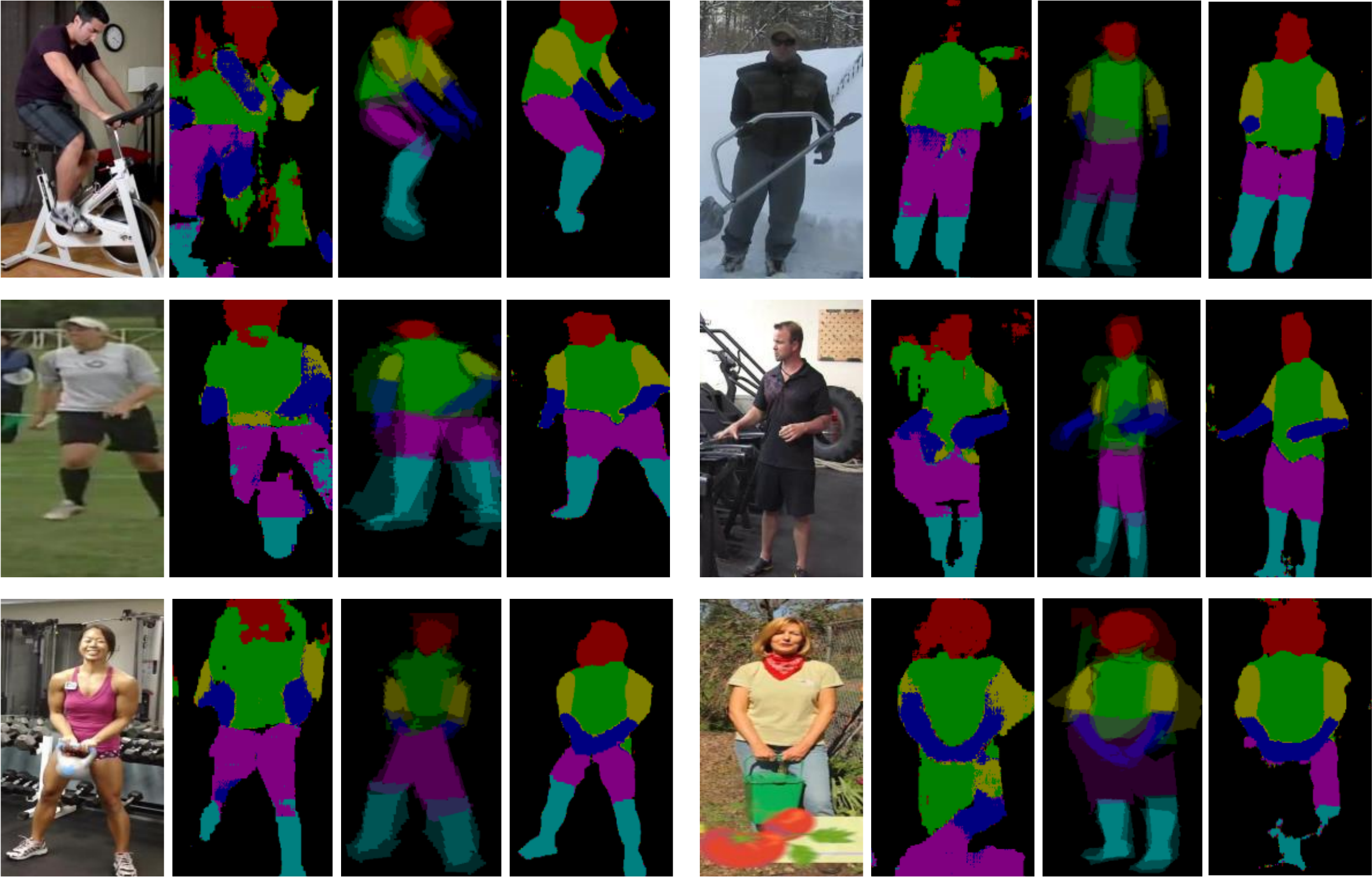}
\end{center}
\vspace{-3mm}
   \caption{Qualitative comparison of the refinement network trained with and without part-level prior. For each image group, from left to right: input image, FCN prediction without prior, corresponding part-level prior for the input image, prediction from refinement network trained with prior.}
\label{fig:prior}
\vspace{-2mm}
\end{figure}

\begin{figure*}[t!]
\begin{center}
\includegraphics[width=0.9\linewidth]{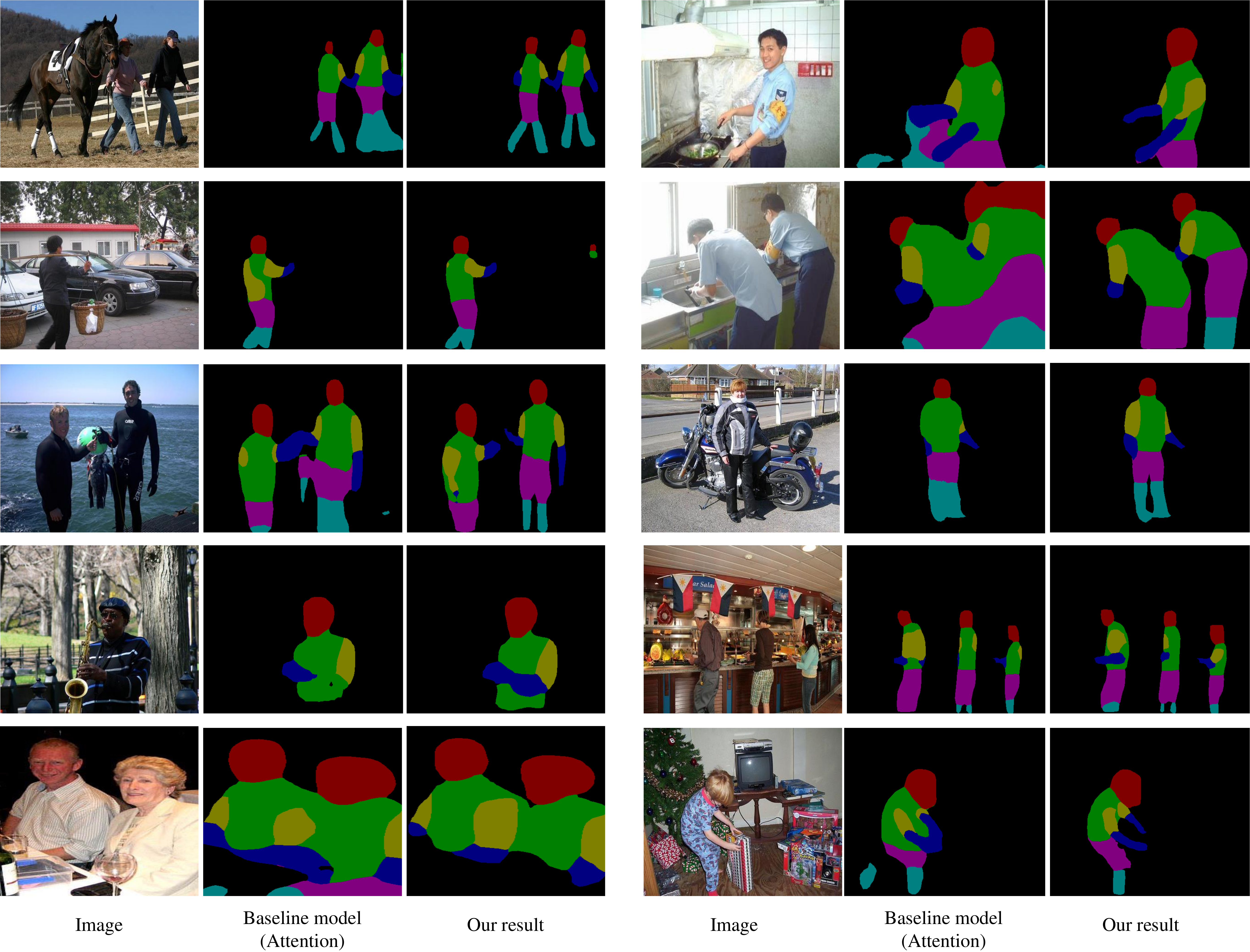}
\end{center}
\vspace{-3mm}
   \caption{Qualitative comparison on the PASCAL-Person-Part dataset between the baseline model and our semi-supervised model.}
\label{fig:res}
\vspace{-3mm}
\end{figure*}
\paragraph{Evaluation of Annotation Number.}
We first compare the performance of the parsing network trained with a different number of annotations on validation set and the results are reported in Table~\ref{tab:annotation}. When all annotations of semantic part segmentations are used, the result is the original implementation of the baseline model~\cite{chen2016attention}, and our reproduction is slightly (0.5 mIOU) higher. Then we gradually add the data with keypoint annotations. As can be seen, the performance increases in line with the number of keypoint annotations. This suggests that our semi supervision method is effective and scalable. Fig.~\ref{fig:res} gives some qualitative comparisons between the predictions of the baseline model and our semi-supervised model.
\begin{table}[t]
	\begin{center}
\begin{small}
		\begin{tabular}{c|c|c|c}
			\hline
			supervision & mask anno.\# & keypoints anno.\# & mIoU\\
			\hline
			\hline
			full & 7K & - & 56.89\\
            \hline
			\multirow{3}{*}{semi} & 7K & 4K & 59.60\\
			     & 7K & 7K & 61.44\\
			     & 7K & 10k  & 62.60\\
            \hline
		\end{tabular}
\end{small}
	\end{center}
\vspace{-3mm}
	\caption{Results on \textbf{PASCAL-Person-Part} dataset. In the ``supervision'' column, ``full'' means all training samples are with segmentation mask annotations, ``semi'' means mixtures of mask annotations and keypoints annotations. }
	\label{tab:annotation}
\vspace{-3mm}
\end{table}

\begin{figure}[t]
\begin{center}
   \includegraphics[width=1\linewidth]{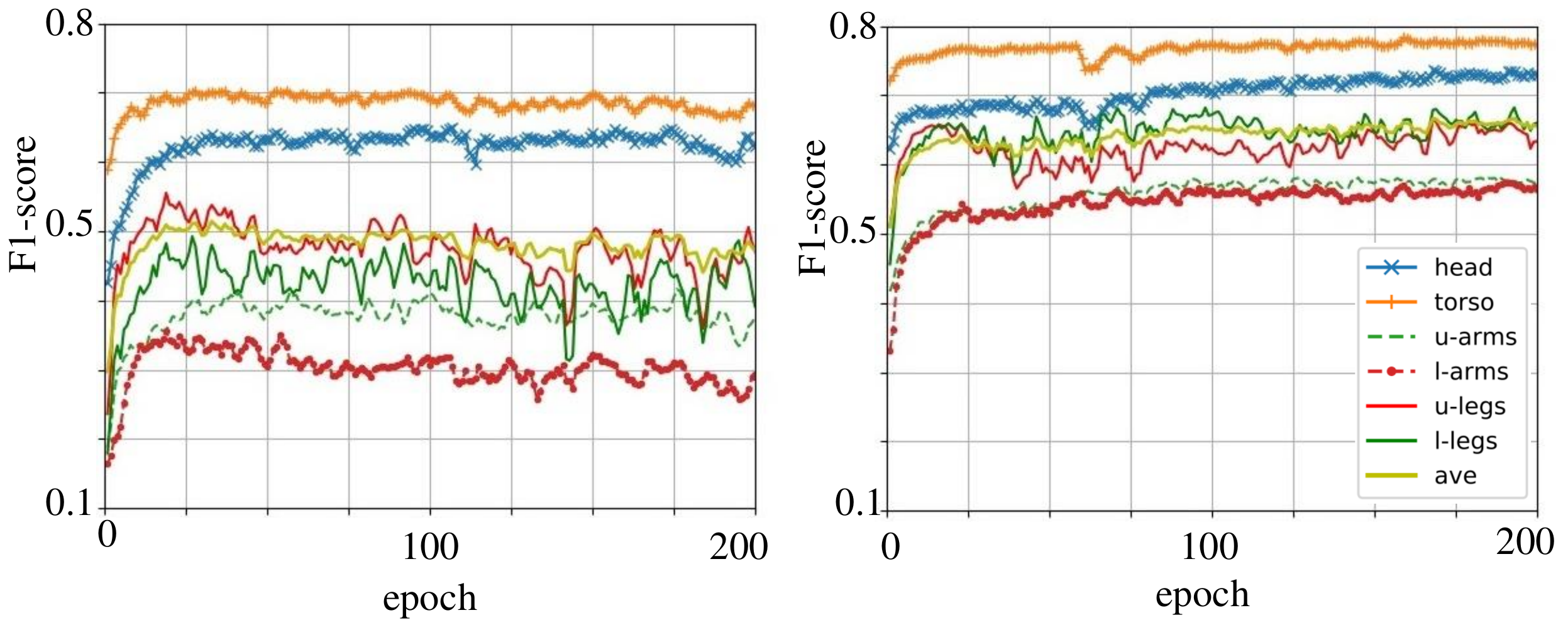}
\end{center}
\vspace{-4mm}
   \caption{Validation accuracy of refinement network with and without part-level prior. The left curve is the case trained without prior and the right one is the case with prior. The performance of refinement network is much better with the part-level prior.}
\label{fig:accuracy}
\vspace{-2mm}
\end{figure}
\vspace{-2mm}
\paragraph{Significance of Part-level Prior.}
We evaluate the significance of the part-level prior. First, we compare the performance of refinement network trained with or without part-level prior on the cropped single person part segmentation validation set. Without prior, the refinement network is a regular FCN trained with limited training data. The validation accuracy for both cases is shown in Fig.~\ref{fig:accuracy}. As we can see, the accuracy of the refinement network with part-level prior is much higher than the case without prior. In Fig.~\ref{fig:prior}, we visualize some predictions from our refinement network trained with and without part-level priors,  and the corresponding priors are also visualized.

\begin{table}[t]
	\begin{center}
\begin{small}
		\begin{tabular}{c|c|c|c}
			\hline
			strategy & cluster size \emph{k} & pool size \emph{n} & mIoU\\
			\hline
			\hline
			skeleton label map~\cite{xia2017joint} & - & - & 58.26\\
            \hline
			\multirow{4}{*}{part-level prior, w/o aug} & 1 & - & 60.10\\
			     & 3 & - & 61.78*\\
			     & 5 & -  & 61.26\\
                 & 7 & -  & 60.32\\
            \hline
            part-level prior, w aug & 3 & 5 & \underline{62.60}\\
            \hline
		\end{tabular}
\end{small}
	\end{center}
	\caption{Performances of different prior generation strategies on \textbf{PASCAL-Person-Part} dataset. }
	\label{tab:strategy}
\vspace{-4mm}
\end{table}
\vspace{-2mm}
\paragraph{Prior Generation Strategy.}
\label{para:strategy}
During our training process, the quality of part-level prior is important and directly affects the refined part segmentation results. We compare different prior generation strategies and report the final results in Table~\ref{tab:strategy}. We first explore using the skeleton label map~\cite{xia2017joint}as prior, which draws a stick with
width 7 between neighboring joints, and the result is 58.26 mIOU. Comparing to this method, our proposed part-level prior has a considerable improvement, which indicates the importance of knowledge transfer during prior generation.

For part-level prior, we compare the impact of the size $k$ of pose-similar cluster. As aforementioned in Sec.~\ref{sec:cluster}, if we only choose the person with the nearest pose and take his/her morphed part parsing result as our prior, the performance is limited. But if we choose too many people to generate our part-level prior, the quality of the final results would also decline. We claim that this is because our part segmentation dataset is small and the intra-cluster part appearance variance would increase as the cluster size increases. Then, we explore the influence of adding data augmentation during the training of refinement network. As we can see, randomly sample 3 candidates to generate part-level prior from a larger pool with size 5 is beneficial for training since it increases the sample variances. For the remaining experiments, we set $k$ as 3 and $n$ as 5.
\vspace{-3mm}
\paragraph{Comparisons with State-of-the-Art.}
In Table~\ref{tab:person}, we report comparisons with the state-of-the-art results on the Pascal-Person-Part dataset~\cite{chen2014detect}. With additional training data generated by our refinement network, our VGG16 based model achieves the state-of-the-art performance.
\begin{table*}[!tp]\setlength{\tabcolsep}{2pt}
	\centering
	\begin{tabular}{cccccccccccccccccccccc}
		\toprule
		{Method} &  head   &  torso  &  u-arms  & l-arms & u-legs & l-legs & Bkg & Avg \\
		\midrule
		DeepLab-LargeFOV-CRF~\cite{chen2015deeplab}  & 80.13 & 55.56 & 36.43 & 38.72 & 35.50 & 30.82 & 93.52 & 52.95 \\
		Attention~\cite{chen2016attention} & 81.47 & 59.06 & 44.15 & 42.50 & 38.28 & 35.62 & 93.65 & 56.39\\
        HAZN~\cite{xia2016zoom} & {80.79} & 80.76 & 45.65 & 43.11 & 41.21 & 37.74 & 93.78 & 57.54\\
		Graph LSTM~\cite{liang2016semantic} & {82.69} & {62.68} & {46.88} & {47.71} & {45.66} & {40.93} & {94.59} & {60.16} \\
		Structure-evolving LSTM~\cite{liang2017interpretable} & 82.89 & 67.15 & 51.42 & 48.72 & 51.72 & 45.91 & 97.18 & 63.57\\
        \midrule
        Joint (VGG-16, +ms)~\cite{xia2017joint} & 80.21 & 61.36 & 47.53 & 43.94 & 41.77 & 38.00 & 93.64  & 58.06\\
		Joint (ResNet-101, +ms)~\cite{xia2017joint} & 85.50 & 67.87 & 54.72 & 54.30 & 48.25 & 44.76  & 95.32  & 64.39\\
        \midrule
        Ours (VGG-16) & 84.06 & 67.03 & 51.66 & 50.15 & 45.33 & 44.26 & 95.73  & 62.60\\
        Ours (ResNet-101) & 84.83 & 68.64 & 53.11 & 53.01 & 48.40 & 46.76  & 95.22  & 64.28\\
        Ours (ResNet-101, +ms) & \textbf{87.15} & \textbf{72.28} & \textbf{57.07} & \textbf{56.21} & \textbf{52.43} & \textbf{50.36}  & \textbf{97.72}  & \textbf{67.60}\\
		\bottomrule
	\end{tabular}%
	\caption{Comparison of semantic object parsing performance with several state-of-the-art methods on the PASCAL-Person-Part dataset~\cite{chen2014detect}. ``+ms'' denotes testing with multi-scale inputs. Note that we only perform single scale testing for our VGG-16 entry since the base network~\cite{chen2016attention} has explicitly utilized multi-scale features.}\label{tab:person}
	\vspace{-4mm}
\end{table*}
Although we focus on exploiting human keypoints annotations as extra supervision, our method applies to other baseline models and can benefit from other parallel improvements. To prove that, we replace our baseline model with ResNet-101~\cite{he2016deep} based DeepLabv2~\cite{chen2015deeplab} model and follow the same training setting. Without additional keypoints annotations, this baseline achieves $59.6$ mAP. Under our semi-supervised training, this model achieves $\mathbf{64.28}$ mIOU. Note that this result is obtained by single scale testing. By performing multi-scale testing (scale = {0.5, 0.75, 1}), we can further achieve $\mathbf{67.6}$ mIOU, which outperforms previous best result by $\mathbf{3}$ mIOU.

\begin{figure}[t]
\begin{center}
%\fbox{\rule{0pt}{2in} \rule{0.9\linewidth}{0pt}}
   \includegraphics[width=0.9\linewidth]{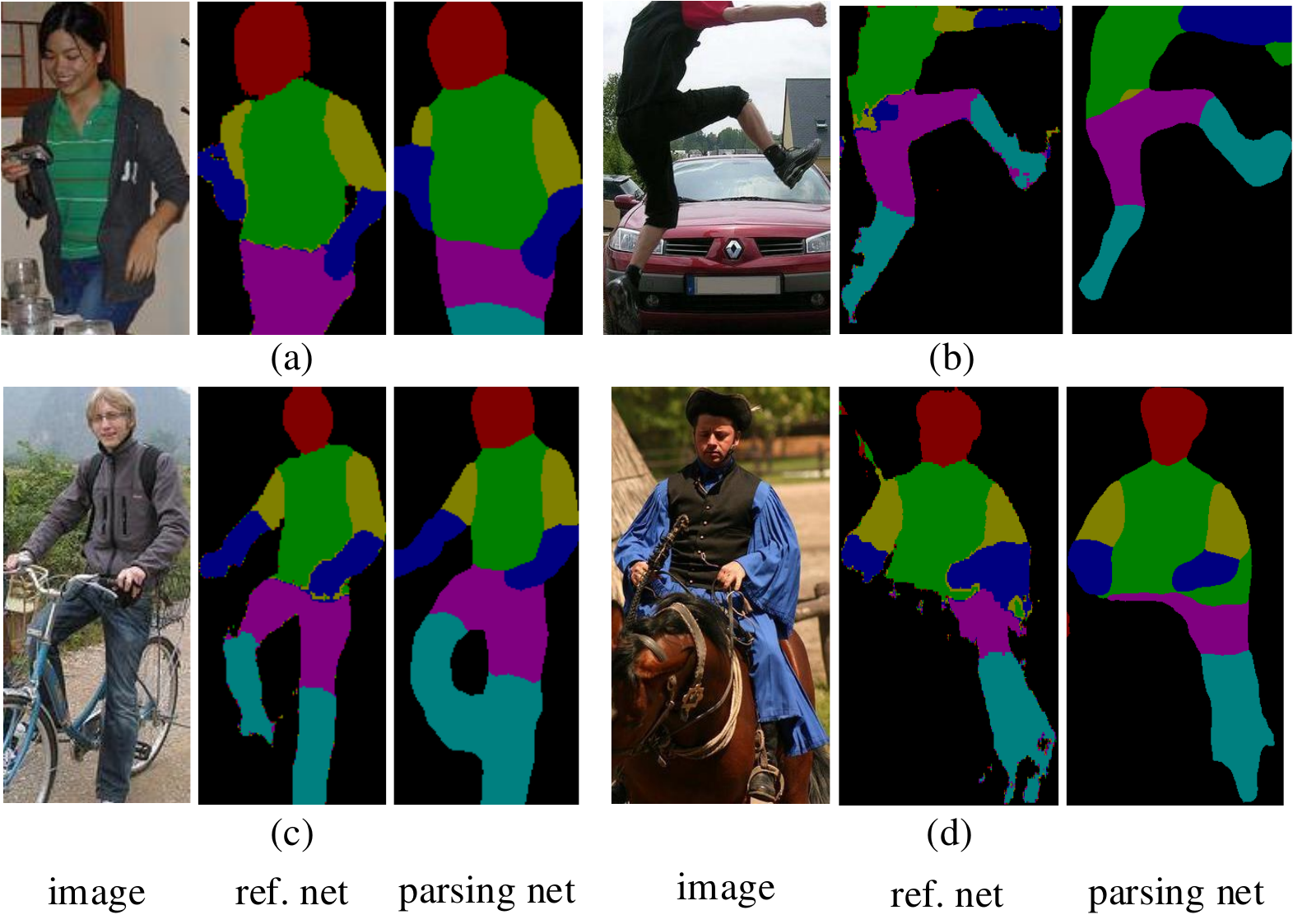}
\end{center}
\vspace{-4mm}
   \caption{Qualitative comparison between the predictions of the refinement network and the parsing network. ``ref.net'' denotes the predictions of the refinement network, and ``parsing net'' denotes the predictions of the parsing network.}
\label{fig:cmp}
\vspace{-5mm}
\end{figure}
\vspace{-5mm}
\paragraph{Comparisons between two Networks.}
To see how the refinement network can assist the training of the parsing network, we visualize some predictions of both networks in Fig.~\ref{fig:cmp}. In this experiment, the parsing network we use has already been trained under the semi supervision setting. As shown in Fig.~\ref{fig:cmp}, due to the guidance of the strong part-level prior, the refinement network makes fewer mistakes on the structure predictions (e.g., the upper legs in (a) and the upper arms in (b)), and produces tighter masks (e.g., the legs in (c)). These improvements obtained by using the part-level prior will be transferred to the parsing network during the semi-supervised training. On the other hand, by leveraging a large number of training data, the parsing network can figure out which are important or irrelevant features and produce predictions that are less noisy (e.g.,(b) and (d)).

\paragraph{Training on Pose Estimation Results.}
Since single person pose estimation is mature enough to be deployed, what if we replace the ground-truth keypoint annotations with pose estimation results? To answer that, we use the pose estimator~\cite{fang2017rmpe} trained on MPII dataset to estimate human poses in COCO dataset~\cite{lin2014microsoft}. Same as previous, we crop those people with full body annotations and collect 10K images. The semi-supervised result achieves $\mathbf{61.8}$ mIOU, which is on par with the results trained on ground-truth annotations. It shows that our system is robust to noise and suggests a promising strategy to substantially improve the performance of human semantic part segmentation without extra costs.
\vspace{-4mm}
\paragraph{Extension to other Categories.}
To show the potential of extending our method to other categories, we also perform experiments on the Horse-Cow Dataset~\cite{wang2015semantic}. The results are reported in Table~\ref{tab:horsecow}. Our baseline model, which is the attention model~\cite{chen2016attention}, has an mIOU of 71.55 for the Horse and an IOU of 68.84 for the Cow. By leveraging the keypoint annotations provided by~\cite{PoseletsPAMI}, we gain improvements of 3.14 and 3.39 mIOU for Horse and Cow respectively. The consistent improvements across different categories indicate that our method is general and applicable to other segmentation tasks where their anatomical similarity can be annotated by keypoints. Finally, by replacing our baseline model with the deeper ResNet-101 based model, we achieve the most state-of-the-art result on the Horse-Cow dataset, yielding the final performances of $76.99$ and $74.22$ mIOU for these two categories respectively.

\begin{table}[!tp]\setlength{\tabcolsep}{3.3pt}
	\centering\scriptsize
	\begin{tabular}{cccccccccccccccccc}
		\toprule
		& & \textbf{Horse} & & &\\
		\hline
		{Method} &  Bkg   &  head  &  body  & leg & tail & Avg \\
		\midrule
		HAZN~\cite{xia2016zoom} & 90.94 & 70.75 & 84.49 & 63.91 & 51.73 & 72.36\\
		{Graph LSTM}~\cite{liang2016semantic} & {91.73} & {72.89} & {86.34} & {69.04} & {53.76} & {74.75}\\
		Structure-evolving LSTM~\cite{liang2017interpretable} & 92.51 & 74.89 & 87.55 & 71.93 & \textbf{57.45} & 76.87\\
		\midrule
        Attention~\cite{chen2016attention} & {90.48} & {68.91} & {83.34} & {64.20} & {50.74} & {71.55}\\
        Ours(VGG-16) & {91.62} & {72.75} & {87.24} & {69.52} & {52.32} & {74.69}\\
        Ours(ResNet-101) & \textbf{93.39} & \textbf{75.63} & \textbf{88.39} & \textbf{72.61} & 54.95 &\textbf{76.99}\\
        \midrule
		& & \textbf{Cow}&  & &\\
		\hline
		{Method} &  Bkg   &  head  &  body  & leg & tail & Avg \\
		\midrule
		HAZN~\cite{xia2016zoom} & 90.71 & 75.18 & 83.33 & 57.42 & 29.37 & {67.20} \\
		{Graph LSTM}~\cite{liang2016semantic} & {91.54} & {73.88} & {85.92} & 63.67 & {35.22} & {70.05}\\
		Structure-evolving LSTM~\cite{liang2017interpretable} & {92.88} & {77.75} & {87.91} & {67.60} & \textbf{42.86} & {73.80}\\
		\midrule
        Attention~\cite{chen2016attention} & {91.06} & {74.33} & {84.38} & {60.60} & {33.85} & {68.84}\\
        Ours(VGG-16) & {92.27} & {78.36} & {88.20} & {64.76} & {37.58} & {72.23}\\
        Ours(ResNet-101) & \textbf{93.70} & \textbf{80.30} & \textbf{89.53} & \textbf{66.92} & 40.65 &\textbf{74.22}\\
		\bottomrule
	\end{tabular}%
\caption{Comparison of object parsing performance with three state-of-the-art methods over the Horse-Cow object parsing dataset~\cite{wang2015semantic}. We also list the performance of Attention~\cite{chen2016attention}, which is the baseline model for our VGG-16 entry.}
\label{tab:horsecow}
\vspace{-4mm}
\end{table}%
%------------------------------------------------------------------------
\vspace{-1mm}
\section{Conclusion}
\vspace{-2mm}
In this paper, we propose a novel strategy to utilize the keypoint annotations to train deep network for human body part parsing. Our method exploits the constraints on biological morphology to transfer the part parsing annotations among different persons with similar poses. Experimental results show that it is effective and general. By utilizing a large number of keypoint annotations, we achieve the most state-of-the-art result on human part segmentation.
\vspace{-3mm}
{\small
\paragraph{Acknowledgement:} This work is supported in part by the National Natural Science Foundation of China under Grants 61772332. This work is also support in part by Sensetime Ltd
}
{\small
\bibliographystyle{ieee}
\bibliography{egbib}
}

\end{document}